\documentclass{article}
\usepackage{graphicx} 
\usepackage{amsmath}
\usepackage{amssymb}
\usepackage{url}
\date{}
\usepackage[a4paper, left=3cm, right=3cm, top=2.5cm, bottom=2.5cm]{geometry}
\usepackage{authblk}  
\usepackage[numbers]{natbib} 
\usepackage[colorlinks=true, linkcolor=blue, citecolor=blue, urlcolor=blue]{hyperref}

\newcommand{\keywords}[1]{%
  \begin{center}
    \textbf{Keywords: } #1
  \end{center}
}

\title{MVG4D: Image Matrix-Based Multi-View and Motion Generation for 4D Content Creation from a Single Image}

\author[1]{Dongfu Yin}
\author[1,2]{Xiaotian Chen\thanks{Corresponding author: chenxiaotian2023@email.szu.edu.cn}}
\author[1]{Fei Richard Yu}
\author[3]{Xuanchen Li}
\author[4]{Xinhao Zhang}

\affil[1]{Guangdong Laboratory of Artificial Intelligence and Digital Economy (Shenzhen)}
\affil[2]{Shenzhen University}
\affil[3]{Tsinghua University}
\affil[4]{Dalian University of Technology}

\begin{document}
\maketitle
    \begin{abstract}
        Advances in generative modeling have significantly enhanced digital content creation, extending from 2D images to complex 3D and 4D scenes. Despite substantial progress, producing high-fidelity and temporally consistent dynamic 4D content remains a challenge.  In this paper, we propose MVG4D, a novel framework that generates dynamic 4D content from a single still image by combining multi-view synthesis with 4D Gaussian Splatting (4D GS). At its core, MVG4D employs an image matrix module that synthesizes temporally coherent and spatially diverse multi-view images, providing rich supervisory signals for downstream 3D and 4D reconstruction. These multi-view images are used to optimize a 3D Gaussian point cloud, which is further extended into the temporal domain via a lightweight deformation network.   Our method effectively enhances temporal consistency, geometric fidelity, and visual realism, addressing key challenges in motion discontinuity and background degradation that affect prior 4D GS-based methods. Extensive experiments on the Objaverse dataset demonstrate that MVG4D outperforms state-of-the-art baselines in CLIP-I, PSNR, FVD, and time efficiency.   Notably, it reduces flickering artifacts and sharpens structural details across views and time, enabling more immersive AR/VR experiences.   MVG4D sets a new direction for efficient and controllable 4D generation from minimal inputs.
    \end{abstract}
    
 
    \vspace{1\baselineskip}
    
    \section{Introduction}
	With the rapid development of generative modeling technologies, the ability to create various digital contents such as 2D images \cite{rombach2022high}, videos \cite{blattmann2023stable,chai2023stablevideo,Esser_2023_ICCV}, and 3D scenes \cite{long2024wonder3d,liu2024oneplus,Chen_2025_CVPR,NEURIPS2022_cebbd24f} has been significantly improved. However, generating continuous and high-quality dynamic 4D scenes \cite{singer2023text,ling2024align} remains challenging. Dynamic 4D scenes are typically represented by Dynamic Neural Radiance Fields(NeRF) \cite{mildenhall2021nerf},which aim to model appearance, geometry, and motion across different viewpoints. For instance, Hexplane \cite{cao2023hexplane} technology enables the generation of dynamic videos directly from text, which can then be converted into 4D presentations through advanced diffusion models, greatly enriching content creation possibilities. Emerging technologies such as CONSISTENT4D \cite{jiang2024consistent4d} and MAV3D \cite{singer2023text}, attempt to generate coherent 4D scenes from a single video source by optimizing the NeRF framework and diffusion models. These methods effectively combine video information with dynamic 3D models to generate 4D content rich in details and dynamic textures. Additionally, techniques such as Animate124 \cite{zhao2023animate124} explore the transformation of a single image into a dynamic 3D video, further expanding the applications of dynamic scene generation.
    
	Despite the potential of these advanced techniques, challenges remain in terms of clarity and time efficiency. Current methods may require long processing times to generate high-quality dynamic NeRF, and the sharpness of the generated dynamic scenes needs improvement. Enhancing efficiency and clarity while ensuring quality remains a key focus of current research in 4D generation technology.
    
	Recent works have introduced time-set variables to optimize Gaussian deformation fields based on the 3D Gaussian Splatting (3D GS) \cite{kerbl20233d} technique, resulting in a novel 4D representation called 4D Gaussian Splatting (4D GS) \cite{wu20244d}. This method aims to achieve efficient training and storage for real-time dynamic scene rendering while maintaining high-quality output. The 4D GS method employs a lightweight MLP to predict 3D Gaussian deformations for new timestamps, offering a novel display representation for dynamic scenes. This approach significantly reduces optimization time and enables real-time processing while maintaining high-quality deformable Gaussian Splatting. 
    
	While 4D Gaussian Splatting (4D GS) offers an efficient method for 4D scene representation and has significantly improved optimization time for 4D content, challenges remain in generating 4D content from a single image using 4D GS technology. These challenges include issues related to the absence of background points and inaccuracies in camera positioning, all of which impede the precise optimization of dynamic scenes. Additionally, during motion, the displacement of Gaussian points can lead to surface tearing on 3D objects, compromising the quality and realism of the rendering. Therefore, the current methods for real-time 4D scene generation based on 4D GS still require further optimization. To address these challenges, we propose a multi-view image generation model that captures comprehensive multi-view information of the input object, addressing issues like missing background points and imprecise camera positioning while significantly reducing the generation time for 4D scenes.
    
	Our work can be summarized as follows:
    
	1. We introduce an image matrix generation module that produces diverse and temporally coherent multi-view images from a single input image. This module provides rich spatial and temporal supervision, significantly enhancing the clarity, consistency, and realism of 4D dynamic content.
    
	2. We develop a compact 3D scene representation by initializing and optimizing 3D Gaussian point clouds directly from the generated image matrix. This representation enables accurate reconstruction with reduced computational overhead, serving as a strong foundation for dynamic modeling.
    
	3. We build an integrated framework that transforms a single image into dynamic 4D content by sequentially combining multi-view generation, 3D reconstruction, and 4D optimization. The proposed method achieves high visual fidelity and temporal consistency while significantly reducing generation time.
	
    \section{Related Work}
	\subsection{Single View to Multi-view Image Generation}
	Exciting progress has been made in using diffusion models to generate images from multiple views using a single image of an object. For example, Zero123 \cite{liu2023zero} utilizes the geometric prior learned by a large-scale diffusion model to train a conditional diffusion model on synthetic datasets, achieving zero-shot synthesis of new perspective images from single-view inputs. By combining images from different perspectives as target images, Zero123++ \cite{shi2023zero123++} significantly overcomes the problem of no correlation between the images generated by Zero123, and improves the quality and consistency of generating consistent multi-perspective images from a single image.  SyncDreamer \cite{liu2023syncdreamer} further optimizes its adaptability to downstream 3D generation tasks on the basis of the previous two, but it still has shortcomings due to the high hardware requirements.
    
	Although significant progress has been made in generating multi-view images from a single input, there are still several drawbacks. The main problem is that it is difficult to maintain the geometric consistency of objects when generating images from other viewpoints. The generated images may exhibit structural and texture inconsistencies. In addition, the problem of possible loss of details is also a key problem to be solved urgently, especially in images with background. Our method addresses these issues by employing a fine-tuned multi-view image generation module. This approach enhances the continuity and clarity of the generated image by minimizing the viewpoint rotation. This method effectively improves the geometric consistency of the objects in the generated images while reducing the loss of details.
    
	\subsection{Image-to-3D Generation} 
	The technology for generating 3D models from images has been rapidly maturing, and many 3D models representation methods based on point clouds \cite{Chen_2025_CVPR,NEURIPS2022_cebbd24f},  meshes, implicit neural representations \cite{Park_2019_CVPR, Chou_2023_ICCV} and Gaussian Splatting have emerged. Most of these methods, which are built from multi-view images, focus on creating a 3D representation of the scene. One-2-3-45++ \cite{liu2024oneplus} combines a 2D diffusion model and a 3D native diffusion model for multi-view conditions to quickly generate 3D meshes through consistent multi-view image generation and 3D reconstruction. However, this traditional diffusion model-based scheme generally has the problem of long optimization time. 3D GS \cite{kerbl20233d} uses 3D Gaussian scene representation and real-time differentiable renderer to achieve real-time rendering of radiation field, which effectively reduces the time required for 3D content reconstruction and achieves high-quality scene representation by optimizing 3D Gaussian attributes and density control. While the task of real-time 3D reconstruction with just one image input has not yet been addressed, LRM \cite{hong2023lrm} uses a Transformer \cite{vaswani2017attention}-based encoder-decoder architecture to predict Neural Radiance Fields (NeRF) directly from a single image for 3D reconstruction. This has significant implications for the task of 3D reconstruction based on a single image.
    
	Current methods for 3D content generation often rely on iterative optimization of the target model using diffusion models. However, this approach can be time-consuming, leading to high computational costs. In contrast, Transformer-based encoder-decoder architectures offer better scalability and efficiency but can produce blurred textures in occluded areas, resulting in visual distortion. Recent innovations using 3D Gaussian Splatting (3D GS) for 3D content representation excel in real-time rendering. Nevertheless, these methods may exhibit artifacts in regions not observed during training. Our approach addresses these challenges by incorporating a multi-view image generation module. This module generates additional views of the input image, thereby reducing artifacts caused by occluded regions. Additionally, we use 3D GS to represent the 3D scene, which effectively reduces computational time and improves optimization.
    
	\subsection{4D Generation} 
	Dynamic scene rendering, also known as 4D generation work, aims to enable real-time rendering of dynamic 3D scenes with efficient training and storage. In this field, neural radiance Fields (NeRF) techniques have been widely used to represent 4D scenes. For example, Neural-3D-Video \cite{li2022neural} uses time-conditioned NeRF and a series of compact latent codes to represent dynamic scenes, and uses a hierarchical training scheme and light importance sampling to significantly improve the training speed and the perceptual quality of generated images. TiNeuVox \cite{fang2022fast} accelerates the optimization of dynamic radiance fields by combining time-aware voxel features and micro-coordinate deformation networks to represent scenes. This work significantly reduces training time and storage costs while maintaining high-quality rendering. However, the robustness of dynamic radiance field reconstruction methods remains a challenge. RoDynRF's \cite{liu2023robust} method for reconstructing camera trajectories and dynamic radiance fields from randomly captured dynamic monocular videos, enabling rendering that focuses on robust dynamic scenes. And MSTH \cite{NEURIPS2023_df311263}, which uses combinations of hash codes to represent dynamic scenes.
    
	With the development of 3D Gaussian Splatting (3D GS), 4D Gaussian Splatting (4D GS) provides an innovative way to render dynamic scenes, which handles dynamic changes through continuous 4D representation. The 3D GS function is manipulated to adapt to temporal and spatial changes, thus enabling real-time rendering and high-resolution output. 4D GS emphasizes computation and storage efficiency, and utilizes a compact network structure to efficiently capture complex dynamic changes while maintaining high-quality rendering effects. Furthermore, DreamGaussian4D \cite{ren2023dreamgaussian4d} uses an image-to-3D framework to fit a static 3D GS function, and then optimalizes the dynamic representation by learning the motion that drives the video. Finally, the 4D GS is exported as an animated mesh sequence, and the texture map is optimized through a video-to-video process. Recently, Stable Video 4D (SV4D) \cite{xie2024sv4d} provides an innovative approach for dynamic 3D content generation that enables 4D generation via multi-frame multi-view coherence. Unlike traditional approaches that rely on independently trained video generation and novel view synthesis models, SV4D employs a unified diffusion model that is able to generate multi-view videos of dynamic 3D objects from a mutil-view video.
    
	Despite these improvements, issues like motion discontinuity and high computation persist. Our approach enhances dynamic rendering with improved image matrix module and efficient 4D GS optimization, yielding better quality and temporal continuity.

    \section{Method}
\begin{figure}[htbp]
		\centering
		\includegraphics[width=\textwidth]{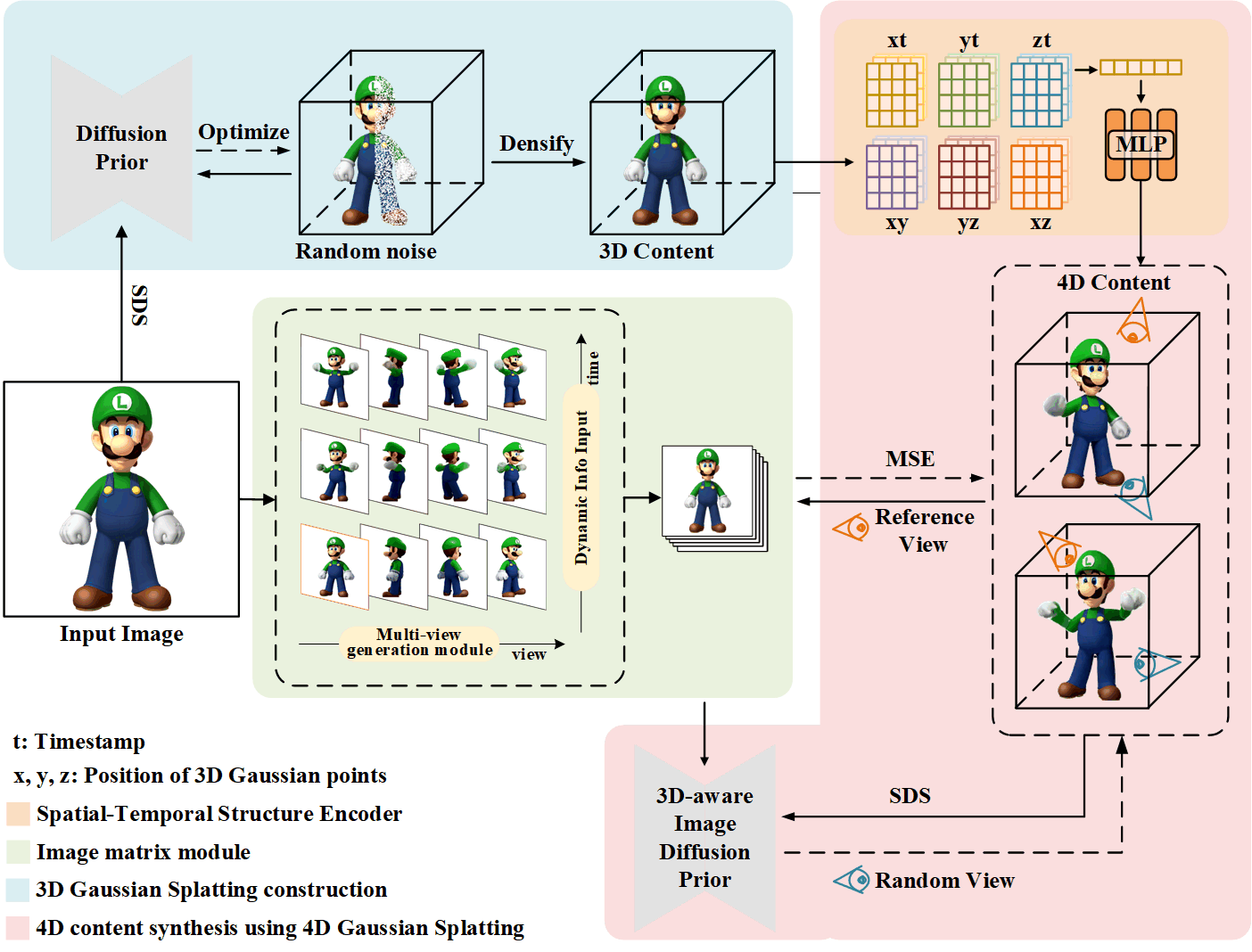}
		\caption{\textbf{The overall pipeline of MVG4D.} MVG4D is divided into three main stages: The green background part is the Image matrix module, the blue background part is the 3D Gaussian Splatting construction, and the red background part is the 4D content synthesis using 4D Gaussian Splatting.}
		\label{fig:1}
	\end{figure}

	The technique for generating 4D content from a single image provides an innovative approach to 4D content generation. Our proposed method, MVG4D, combines a image matrix module with the 4D Gaussian Splatting (4D GS) dynamic content optimization technique. This method is based on 3D Gaussian Splatting (3D GS)  technology to optimize the Gaussian deformation field, which not only enhances the continuity and clarity of the generated 4D content but also significantly accelerates the 4D content generation process. Our study is illustrated in Figure \ref{fig:1}, which outlines three main phases: Image matrix module, 3D Gaussian Splatting Construction, and 4D Content Synthesis Using 4D Gaussian Splatting.

	\subsection{Image matrix module}
	To construct a comprehensive and time-aware multi-view image matrix from a single input image $I_0$, we propose a two-stage processing flow that takes full advantage of the multi-view generation as well as the introduction of dynamic information.
    
	Given the input image $I_0$, we first utilize the pre-trained module to synthesize a video that encapsulates dynamic information. This video is then decomposed into a set of video frames ${I_t}$, where $t$ denotes the timestamp. Although this video captures temporal dynamics, it contains only single-view information. To achieve multi-view dynamic content generation, we fine-tune a diffusion-based multi-view image generation model to synthesize novel viewpoints for each frame ${I_t}$. The resulting image matrix embeds both temporal and viewpoint diversity, serving as a supervisory signal for optimizing the subsequent 4D GS representation.
    
	The core of our approach lies in fine-tuning a view-conditional 2D diffusion model to generate consistent multi-view images for each video frame. During inference, the model takes as input the original frame image and the desired relative camera parameters such as angular offsets and depth variations—and outputs novel-view images accordingly. During training, we place the object at the origin of a canonical 3D coordinate system and simulate a spherical camera setup. The camera is positioned on a sphere centered at the object and is constrained to always face the origin. Let two camera viewpoints be defined by spherical coordinate ($\theta_1$, $\phi_1$, $r_1$) and ($\theta_2$, $\phi_2$, $r_2$), representing the polar angle, azimuth angle, and radius, respectively. Their relative transformation is parameterized as $(\Delta \theta, \Delta \phi, \Delta r) = (\theta_2 - \theta_1, \phi_2 - \phi_1, r_2 - r_1)$.
    
\begin{figure}[ht]
		\centering
		\includegraphics[height=6cm,width=0.75\linewidth]{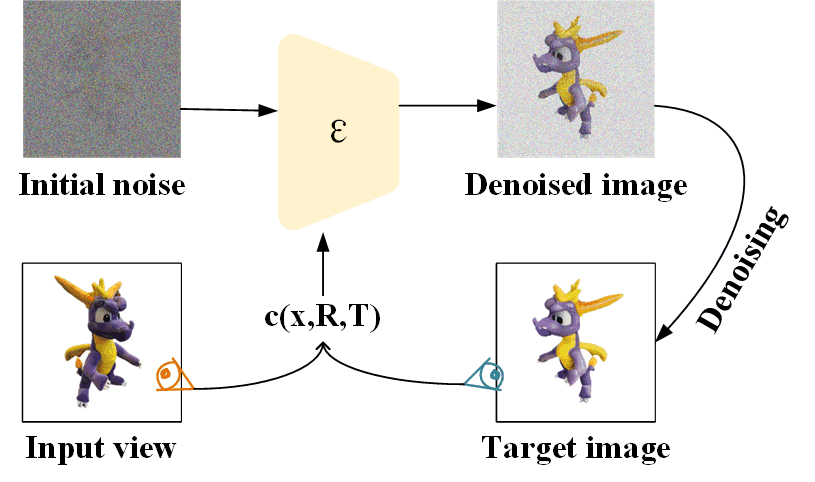}
		\caption{\textbf{The learning process of the multi-view image generation model.} The input image, along with the target view information, is provided as input to a fine-tuned diffusion model, which then generates the corresponding target image.}
		\label{fig:2}
	\end{figure}

	The training objective of the diffusion model is to learn a function $f$ such that given an input image $x_1$ and a viewpoint transformation $(\Delta \theta, \Delta \phi, \Delta r)$, the model can generate a novel-view image that closely resembles the ground truth image $x_2$, which is captured from the target viewpoint. This process is illustrated in Figure~\ref{fig:2}. The model thus learns a general mechanism for camera viewpoint control and can infer the target image $x_2$ from $x_1$ under arbitrary relative viewpoint changes.
    
	The training objective is expressed as:
	\begin{equation}
		\underset{\theta}{\min} \, \, \mathbb{E}_{z \sim \mathcal{E}(x), t, \epsilon \sim N(0,1)} \left\Vert \epsilon - \epsilon_{\theta} (z_t, t, c(x, R, T)) \right\Vert^2_2
	\end{equation}
    
	In this formula, $x$ denotes the input image, $c(x,R,T)$ represents the conditioning embedding that incorporates the input image and the target viewpoint information, $t$ is the diffusion timestep, $\mathcal{E}$ is the image encoder, $\epsilon_{\theta}$ is a U-Net-based denoiser, and $z_t$ is the latent representation of $x$ at timestep $t$.

	Due to the use of different relative view parameters for training, the novel-view images generated from each frame $I_t$ may exhibit inconsistent viewpoint alignments. This inconsistency introduces mismatches in the constructed image matrix, adversely affecting the quality of 4D content generation. To resolve this issue, we further fine-tune the diffusion model using carefully aligned supervision to ensure consistency across views and time. Specifically, during training, we minimize perceptual discrepancies between the generated novel-view images and the corresponding input frames $I_t$, enforcing visual alignment between the synthesized views and the original inputs. This enhancement enables us to generate a spatiotemporally coherent image matrix, thereby improving the downstream optimization of 4D GS.
    
	\subsection{3D Gaussian Splatting Construction}
	After obtaining continuous and high-quality multi-view images, we utilize advanced 3D reconstruction technology to transform these 2D images into a 3D GS model. This process consists of two key steps.
    
	\subsubsection{Multi-view Fusion}
	We employ multi-view fusion technology to integrate image information captured from various perspectives. This technique leverages the geometric and illumination consistency between images to accurately reconstruct the 3D structure.
    
	\subsubsection{Volumetric Reconstruction}
	We initialize unity-scaled, rotation-free 3D Gaussian point clouds at random positions within the space periodically densifying them during optimization.  Unlike the reconstruction pipeline, we begin with fewer Gaussian point clouds but densify them more frequently to align with the generation process.  We optimize the 3D Gaussian point clouds using Score Distillation Sampling (SDS) loss.  At each step, we sample random camera poses centered around the object and render the RGB image from the current viewpoint.  During training, we linearly decrease the time step $t$, which weights the random noise added to the rendered RGB image.  The input image then serves as a 2D diffusion prior, which is used to optimize the underlying 3D Gaussian point clouds with the SDS loss:

	\begin{equation}
		\nabla_{\Theta} \mathcal{L}_{SDS} = \mathbb{E}_{t, p, \epsilon} \left[ \omega_{(t)} \left( \epsilon_\phi (I; t, I^r, \triangle p) - \epsilon \right) \frac{\partial I}{\partial \Theta} \right]
	\end{equation}
	
	Where $\omega_{(t)}$ is the weighting function, $\epsilon_{\phi}$ represents the noise predicted by the 2D diffusion prior $\phi$, $\triangle p$ is the relative change in camera pose with respect to the reference camera $r$. $I$ is the input image, and $I^r$ is the image of the new pose obtained by the 2D diffusion model. Through this method, we can effectively recover the 3D geometry and texture information from 2D images, providing a solid foundation for the subsequent generation of 4D dynamic models. The optimization goal of this stage is to enhance the geometric accuracy and visual-realism of the 3D model. Additionally, it aims to ensure the model's consistency when viewed from various angles.
    
	\subsection{4D Content Synthesis Using 4D Gaussian Splatting}
	After obtaining the image matrix containing multi-view dynamic information of the input image and the initialized 3D GS model, we convert the static 3D GS point cloud into a dynamic 4D model. We extract the central coordinates $(x, y, z)$ and timestamp $t$ of each 3D Gaussian point cloud. A Spatial-Temporal Structure Encoder is then utilized to calculate the characteristics of the voxels. These features are analyzed and decoded using a micro-MLP to obtain the deformed 3D Gaussian point cloud at timestamp $t$, render the deformed point cloud image, and match the image matrix generated in the first stage.  The mean squared error (MSE) loss is calculated between the image matrix generated in the first stage and the images rendered by the deformed point cloud:
	
	\begin{equation}
		\mathcal{L}_{\text{Ref}} = \frac{1}{\mathcal{T}} \sum_{\tau=1}^{\mathcal{T}} \left\Vert f(\phi(S, \tau), o_{\text{Ref}}) - I^{\tau}_{\text{Ref}} \right\Vert^2_2
	\end{equation}
	
	In this formula, $\tau$ is a time variable, $o_{\text{Ref}}$ denotes the viewpoint information corresponding to the image matrix generated in the first stage, with the associated image represented as $I^{\tau}_{\text{Ref}}$. The 4D deformation field model $f$ is used to render the image from this viewpoint, and the error between the rendered image and $I^{\tau}_{\text{Ref}}$ is calculated to optimize the model $f$. To comprehensively address the challenge of modeling occluded parts of the scene, we utilize a 3D-aware image diffusion model based on the multi-view images generated in the initial stage. This approach enables us to capture more complete and detailed scene information, which is then used to reverse-optimize the 4D Gaussian deformation field through SDS loss.
    
	Through our research on the proposed method MVG4G, we can not only reconstruct accurate 3D models from a single image but also generate dynamic 4D content. Compared to traditional methods, our technology significantly improves generation efficiency, the naturalness of dynamic content, and visual continuity. The optimized model also demonstrates enhanced clarity and reduced motion flicker in dynamic performances. These improvements are crucial for the practical application of 4D content in fields such as augmented reality and virtual reality, particularly for creating more realistic and immersive experiences.
    	
    \section{Experiment}
	\subsection{Experiment Setup}
	\subsubsection{Dataset}
	To evaluate the effectiveness of our method in generating high-quality dynamic 4D content, we conducted qualitative and qualitative evaluations based on datasets such as Objaverse \cite{Deitke_2023_CVPR}.
    
	\subsubsection{Evaluation Metrics}
	For the task of generating 4D content from a single input image, designing appropriate evaluation metrics is essential to comprehensively assess the quality, consistency, and realism of the generated multi-view and dynamic content. Our method is evaluated using the following metrics: CLIP-I, PSNR, Time, FVD \cite{unterthiner2018towards}, each targeting different aspects of performance:
    
	\begin{itemize}
		\item CLIP-I: Measures perceptual similarity between the generated image and the input image by computing their cosine similarity in the CLIP embedding space.
		\item PSNR: Evaluates pixel-level reconstruction fidelity between the generated image and the input image; higher values indicate better low-level similarity.
		\item Time: Reports the inference time required to generate the full 4D content from a single image.
		\item FVD-F: calculate FVD over frames at each view. 
		\item FVD-Diag: calculate FVD over the diagonal images of the image matrix. 
		\item FV4D: calculate FVD over all images by scanning them in a bidirectional raster order.
	\end{itemize}
    
	\subsubsection{Baselines}
	To evaluate the quality of dynamic 4D content generation from a single image, we compare our method MVG4D with several recent 4D content generation methods, including DreamGaussian4D \cite{ren2023dreamgaussian4d}, V4D \cite{gan2023v4d}, 4Diffusion \cite{zhang20244diffusion}, etc. We use official code published by the respective authors to generate comparison results.
    
	\subsection{Implementation Details}
	In all of our experiments, we only used a single NVIDIA RTX 4090 GPU. We process a single image of the input using a fine-tuned diffusion model to obtain image matrix supervision that optimizes 4D GS to generate dynamic 4D content.	

        \begin{table}[htbp]
		\renewcommand{\arraystretch}{1.1}
		\centering
		\caption{\textbf{CLIP Image Similarity (CLIP-I).}  CLIP-I assesses the similarity between the input image and the generated 4D rendered image by computing their cosine similarity. The normalized similarity values range from 0 to 1, with higher values indicating that the generated 4D content is more similar to the input image. Compared with current mainstream and novel 4D scene generation methods, the 4D rendered images produced by MVG4D exhibit greater similarity to the input images, highlighting the advantages of our approach in 4D content generation.} 
		\begin{tabular*}{\linewidth}{@{\extracolsep\fill}ll}
			\hline
			Method          & CLIP-I $\uparrow$\\
			\hline
			RealFusion-V \cite{melas2023realfusion}    & 0.803 \\
			Animate124 \cite{zhao2023animate124}      & 0.854 \\
			DreamGaussian4D \cite{ren2023dreamgaussian4d} & 0.923 \\
			Consistent4D \cite{jiang2024consistent4d} & 0.921 \\
			EG4D \cite{sun2024eg4d} & 0.954 \\
			MVG4D (ours)            & \textbf{0.982} \\
			\hline
		\end{tabular*}
		\label{tab:1}
	\end{table}
    
	\subsection{Quantitative Results}
	\subsubsection{Image Similarity Metrics}
	The experimental results, as shown in Table \ref{tab:1}, demonstrate that the 4D rendered images generated by the MVG4D method exhibit significantly higher CLIP-I similarity to the input image compared to those produced by existing mainstream and novel 4D scene generation methods. This suggests that our approach has made substantial progress in enhancing image similarity. The improved performance is likely due to the use of multi-view images, which are generated by our multi-view image generation module. These images provide additional spatial information that effectively guides the optimization of the 4D GS method. This innovation enables us to generate more accurate 4D content compared to methods that rely on video frames.

        \begin{table}[htbp]
		\renewcommand{\arraystretch}{1.1}
		\centering
		\caption{\textbf{Peak Signal-to-Noise Ratio (PSNR).} PSNR is used to assess the quality of 4D scenes generated from a single input image. Our method is benchmarked against several existing methods, showing a significant improvement in PSNR values. Compared to the baseline methods, our approach significantly reduces visual distortion in the reconstructed scene, achieving higher fidelity and more precise generation of dynamic 4D content.} 
		\begin{tabular*}{\linewidth}{@{\extracolsep\fill}ll}
			\hline
			Method          & PSNR $\uparrow$\\
			\hline
			TiNeuVox-B \cite{fang2022fast} & 32.67 \\
			DreamGaussian4D \cite{ren2023dreamgaussian4d} & 34.05 \\
			KPlanes \cite{kplanes_2023} & 31.61 \\
			V4D \cite{gan2023v4d} & 33.72 \\
			4Diffusion \cite{zhang20244diffusion} & 35.07 \\
			MVG4D (ours)            & \textbf{36.44}\\
			\hline
		\end{tabular*}
		\label{tab:2}
	\end{table}    
	
	As shown in Table \ref{tab:2}, when compared with state-of-the-art 4D scene generation methods, the 4D rendering results generated by MVG4D achieved significantly higher PSNR scores, highlighting the effectiveness and advancement of the proposed method. This advantage is likely due to our use of 4D Gaussian point clouds to represent 4D scenes, which better accumulates spatial point information compared to traditional NeRF-based methods.

        \begin{table}[htbp]
		\renewcommand{\arraystretch}{1.1}
		\centering
		\caption{\textbf{Evaluation of 4D outputs on FVD.} The superior FVD results are attributed to the use of 3D Gaussian Splatting, which provides a compact and expressive spatial representation that enables more stable and coherent temporal modeling.}
		\begin{tabular}{llll}
			\hline
			Method          & FVD-F $\downarrow$   & FVD-Diag $\downarrow$ & FV4D $\downarrow$     \\ \hline
			Consistent4D \cite{jiang2024consistent4d}    & 1133.93  & 741.52    & 871.95    \\
			SV3D \cite{voleti2025sv3d}           & 989.53   & 526.78    & 690.49    \\
			STAG4D \cite{zeng2025stag4d}          & 861.88   & 636.83    & 546.56    \\
			SV4D \cite{xie2024sv4d}            & 677.68   & 525.65    & 614.35    \\
			Dreamgaussian4D \cite{ren2023dreamgaussian4d} & 697.8    & 615.68    & 638.15    \\ \hline
			MVG4D(ours)     & \textbf{241.99} & \textbf{201.71} & \textbf{134.58} \\
			\hline
		\end{tabular}
		\label{tab:3}
	\end{table}
    
	As shown in Table \ref{tab:3}, when compared with state-of-the-art 4D scene generation methods, the 4D rendering results produced by MVG4D achieved significantly lower FVD scores across FVD-F, FVD-Diag, and FV4D, indicating better temporal and view consistency in the generated dynamic content. This improvement is largely attributed to our use of 4D Gaussian point cloud.This is largely due to the use of 3D Gaussian Splatting, which provides a compact yet expressive initialization that better captures spatial structure and supports efficient temporal deformation.

        \begin{table}[htbp]
		\renewcommand{\arraystretch}{1.1}
		\centering
		\caption{\textbf{Time efficiency.} Benchmarking based on the Objaverse normalized dataset evaluated the processing time required to generate 4D scenes from input images.  Compared with the existing methods, the proposed method significantly reduces the processing time, which verifies its efficiency.}
		\begin{tabular*}{\linewidth}{@{\extracolsep\fill}ll}
			\hline
			Method          & Time $\downarrow$ \\
			\hline
			V4D \cite{gan2023v4d}          & 6.9 h \\
			TiNeuVox-B \cite{fang2022fast}    & 28 mins \\
			DreamGaussian4D \cite{ren2023dreamgaussian4d} & 13 mins \\
			Stag4D \cite{zeng2025stag4d} & 9 mins  \\
			MVG4D (ours)          & \textbf{8 m 46 s} \\
			\hline
		\end{tabular*}
		\label{tab:4}
	\end{table}
    
	\subsubsection{Time Efficiency Analysis}
	Time is a crucial metric for evaluating the practicality of 4D content generation methods, particularly in real-time applications like AR/VR. We benchmark the total time from a single input image to the successful generation of 4D content. As shown in Table \ref{tab:4}, our method significantly outperforms prior works in generation speed, benefiting from the use of 4D Gaussian Splatting. Compared to NeRF-based approaches, our lightweight deformation network enables efficient timestamp generation from a static 3D Gaussian point cloud, reducing computational overhead.
	
	\subsection{Qualitative Results}
\begin{figure}[htbp]
		\centering
		\includegraphics[width=\textwidth]{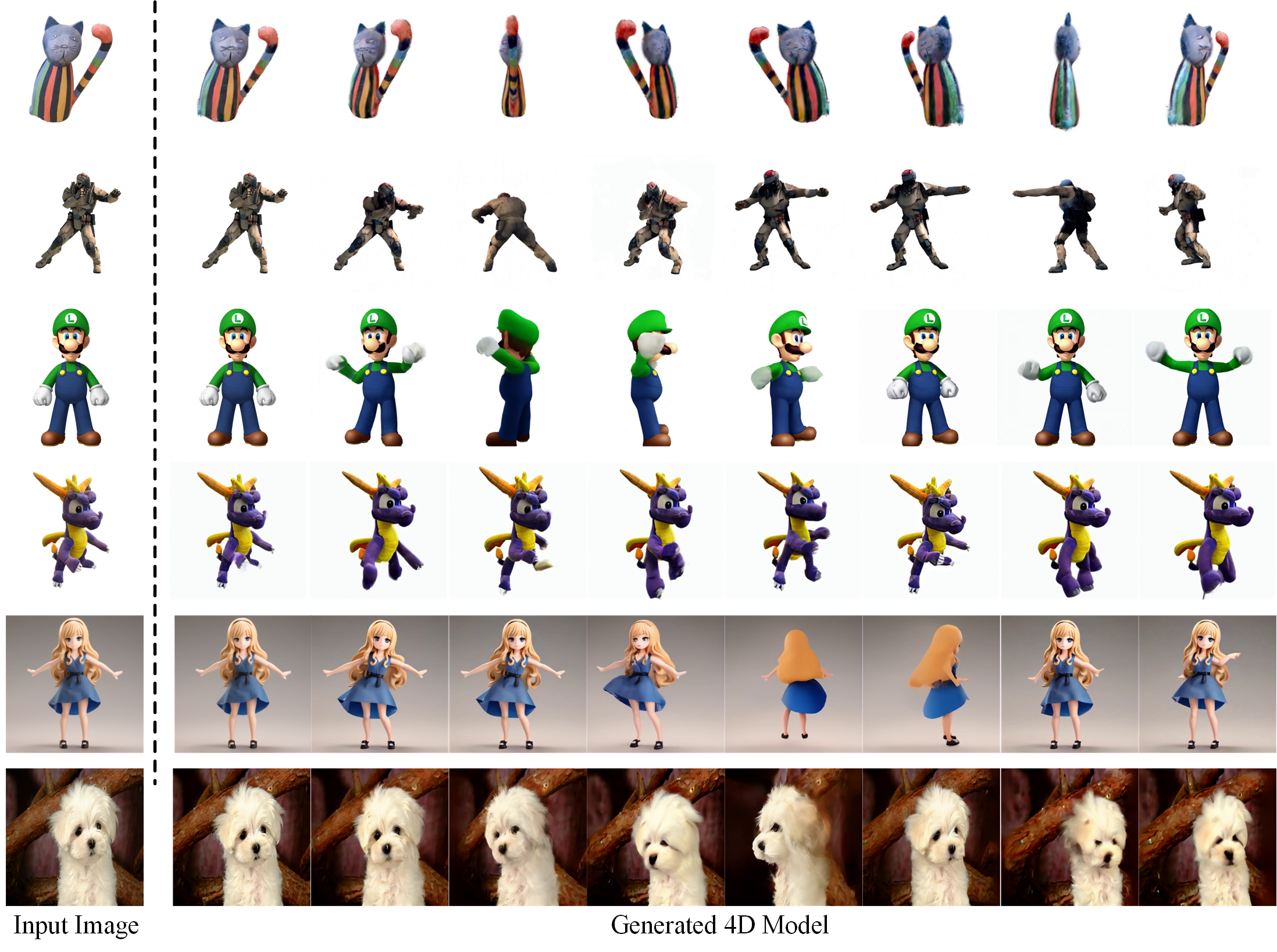}
		\caption{\textbf{Visual representation of the generated results.} In the figure, the rendered images at different timestamps and viewpoints are visually compared with the reference image.}
		\label{fig:3}
	\end{figure}

	To evaluate the effectiveness of our method in generating high-quality dynamic 4D content, we conducted qualitative evaluations based on five examples provided by datasets such as Objaverse.  As depicted in Figure \ref{fig:3}, we provide a visual comparison between a rendered view and a reference image based on a 4D scene representation obtained by 4D GS.  From a qualitative perspective, the generated dynamic 4D scenes maintain a consistent appearance and geometric structure across different viewpoints, demonstrating the exceptional capability of MVG4D to handle spatial transformations.  Additionally, the generated scenes excel in terms of texture detail and motion realism, thereby showcasing the capability of our proposed method to produce high-quality dynamic content.

\begin{figure}[htbp]
		\centering
		\includegraphics[width=\linewidth]{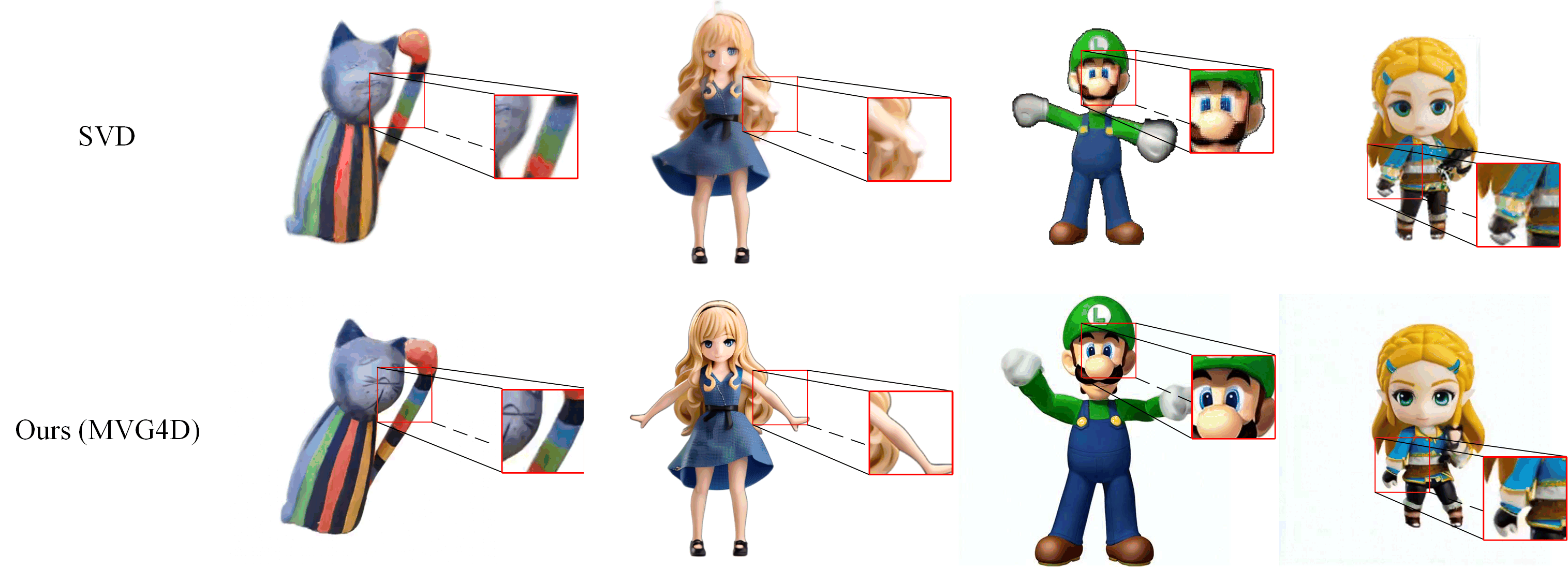}
		\caption{\textbf{Detail enlarge comparison diagram.} This figure illustrates the comparative performance of the baseline method (SVD) and our proposed method (MVG4D) in terms of detail recovery and boundary treatment.}
		\label{fig:4}
	\end{figure}

	Accurate detail rendering and boundary handling are crucial for achieving high-quality dynamic 4D content generation. As shown in Figure \ref{fig:4}, we compare the performance of the benchmark Stable Video Diffusion (SVD) \cite{blattmann2023stable} method and our proposed MVG4D method. The SVD method exhibits noticeable blurring in high-frequency details, with significant noise and blurred transitions along edges, reducing clarity. In contrast, MVG4D achieves sharper boundaries by preserving structural details and effectively suppressing noise. For boundary handling, the SVD method produces a sawtooth effect along edges, resulting in uneven contours and abrupt transitions, whereas MVG4D delivers smoother and more natural results, enhancing overall visual fidelity.

	These results show that MVG4D significantly outperforms current state-of-the-art methods in detail preservation and boundary handling, leading to more realistic and visually compelling 4D dynamic content. This performance gain can be attributed to the use of 3D Gaussian point clouds, which offer richer spatial information than the interval-based sampling strategy in NeRF, thereby enabling higher-quality reconstruction.
    
	\subsection{Ablation Study}
	\subsubsection{Ablation Experiments on Viewpoint Adjustment}
\begin{figure}[htbp]
		\centering
		\includegraphics[width=\textwidth]{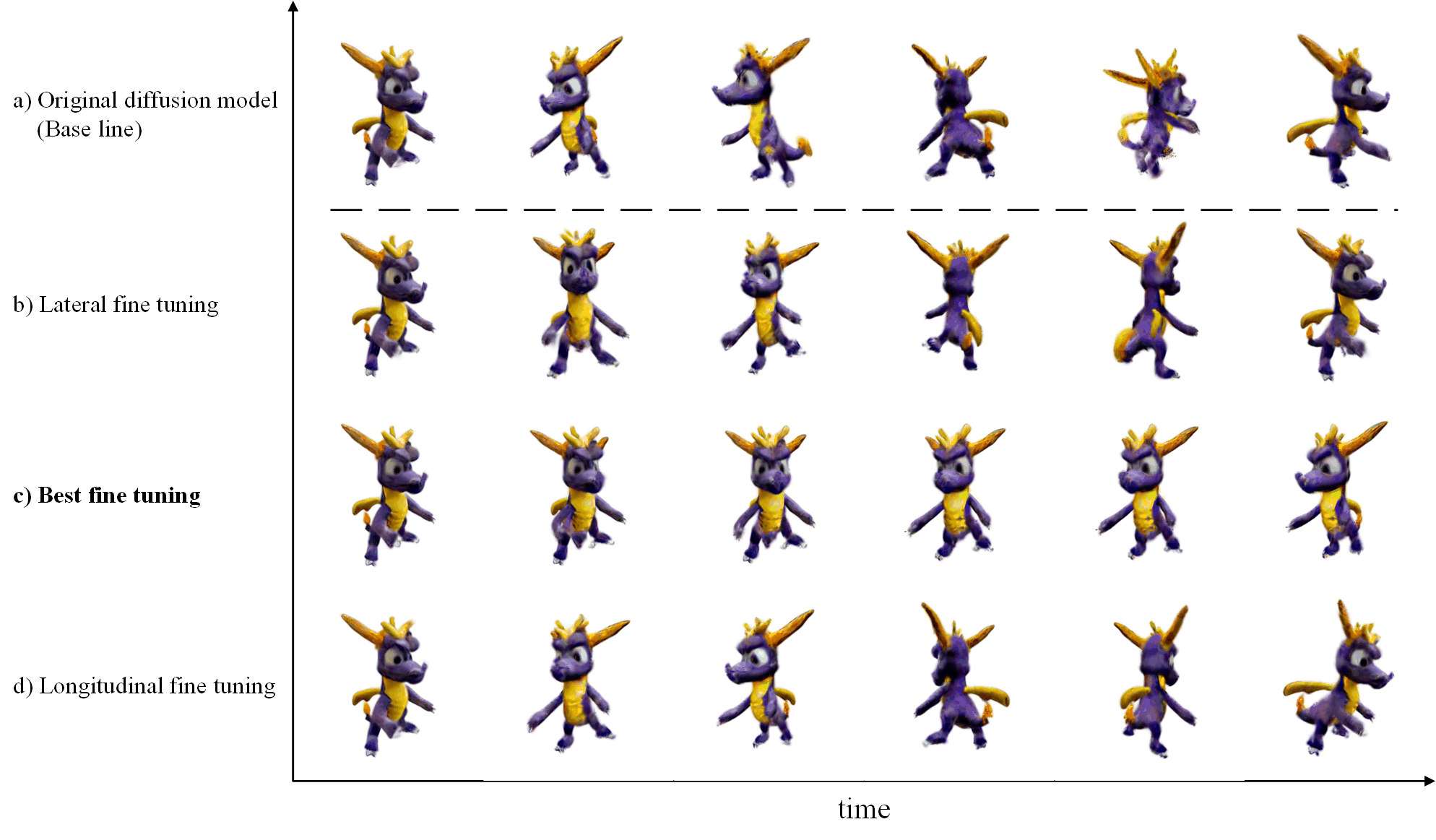}
		\caption{\textbf{Impact of multi-view image generation module on model performance.} Under identical input images and frame rates, the best results were achieved by the version fine-tuned along the vertical axis, which exhibited smoother transitions and reduced motion artifacts.  In contrast, the original diffusion model suffered from flickering effects and abrupt motion changes, highlighting the critical role of the multi-view generation module in achieving high-quality, temporally coherent 4D content.}
		\label{fig:5}
	\end{figure}

	We refer to the enhanced diffusion model used in this study as MVIG (Multi-View Image Generation module). Figure \ref{fig:5} shows the results of an ablation study conducted to evaluate the effectiveness of the viewpoint adjustment in improving the fluency and clarity of generated 4D content and enhancing the original diffusion model. In this study, we froze the 3D content generation and 4D content optimization phases to isolate the impact of the subsequent 4D expression. We compared the original diffusion model with three models fine-tuned separately for vertical and horizontal views, all applied to generate dynamic 4D content from the same input image. Our fine-tuned model significantly outperforms the original diffusion model, demonstrating superior action continuity and fluency, with video frames exhibiting less flickering compared to the original model. This improvement, which maintains the temporal consistency of dynamic 4D content, can be attributed to the multi-view image generation module, which controls fluctuations in view parameters and provides more consistent view information for optimizing the 4D GS model.

        \begin{table}[htbp]
		\renewcommand{\arraystretch}{1.3}
		\centering
		\caption{\textbf{Ablation experiment of MVIG.} The clarity and PSNR of the generated content are lower when the multi-view image generation (MVIG) module is removed compared to when it is enabled, which demonstrates the effectiveness and necessity of the proposed method.}
		\begin{tabular*}{\linewidth}{@{\extracolsep\fill}lll}
			\hline
			\multicolumn{1}{l}{} & CLIP-I $\uparrow$ & PSNR $\uparrow$   \\ \hline
			w/o MVIG             & 0.859  & 30.54 \\
			Half-MVIG            & 0.918  & 32.87  \\
			Full-MVIG            & \textbf{0.982}  & \textbf{36.44}  \\ \hline
		\end{tabular*}
		\label{tab:5}
	\end{table}
    
	\subsubsection{Ablation Experiments on MVIG}
	Table \ref{tab:5} demonstrates the significance of our Multi-View Image Generation module(MVIG) in improving the quality of 4D content. We denote the method without MVIG, which directly uses the input single image for supervised 4D Gaussian Splatting (4D GS) optimization, as "w/o MVIG". The method using only the first stage of the MVIG process is labeled "Half-MVIG". Experimental results indicate that the full MVIG, referred to as "Full-MVIG", significantly enhances all evaluation metrics. Specifically, "Full-MVIG" achieves the highest CLIP-I and PSNR scores, indicating that it produces the most semantically consistent and high-definition 4D content.

    \section{Conclusion}
	We propose MVG4D, a novel framework for efficiently generating dynamic 4D content from a single static image. Central to our method is an image matrix module that synthesizes a temporally coherent and spatially diverse set of multi-view images, providing dense supervision for subsequent 4D representation learning. By integrating this module with dynamic content optimization, our approach significantly improves the temporal continuity, geometric consistency, and visual clarity of the rendered 4D scenes. Experimental results demonstrate that MVG4D excels at reconstructing dynamic content with high fidelity and realism, closely aligning with the appearance of the input image. These advancements lay a strong technical foundation for dynamic 4D content applications in areas such as augmented reality and virtual reality. Future work may explore more complex scenarios, including background control and scene composition, to further enhance the versatility and expressiveness of our framework.

    \newpage 
    \bibliography{references}

\end{document}